\documentclass[runningheads]{llncs}
\usepackage{amsmath}
\usepackage{cleveref}
\usepackage{booktabs} 
\usepackage{caption} 
\usepackage{subcaption} 
\usepackage{graphicx}
\usepackage{pgfplots}
\usepackage{amssymb}
\usepackage[all]{nowidow}
\usepackage[utf8]{inputenc}
\usepackage{tikz}
\usetikzlibrary{er,positioning,bayesnet}
\usepackage{multicol}
\usepackage{algpseudocode,algorithm,algorithmicx}

\definecolor{blue}{HTML}{1F77B4}
\definecolor{orange}{HTML}{FF7F0E}
\definecolor{green}{HTML}{2CA02C}

\pgfplotsset{compat=1.14}

\begin{document}
\title{Reverb: Open-Source ASR and Diarization from Rev}
%
%
\author{Nishchal Bhandari \and
Danny Chen \and
Miguel \'Angel del R\'io Fern\'andez\and
Natalie Delworth \and
Jennifer Drexler Fox \and
Mig\"uel Jett\'e \and
Quinten McNamara \and
Corey Miller \and
Ond\v rej Novotn\'y \and
J\'an Profant \and
Nan Qin \and
Martin Ratajczak \and
Jean-Philippe Robichaud}
%
%
\institute{Rev}
\maketitle              
%
%
%
%
\section{Introduction}
Rev, as a leader in human transcription of English, has amassed the largest high quality English speech recognition dataset in the world. The research team at Rev has used this corpus to develop extremely accurate speech recognition and speaker diarization models, currently available through the rev.ai\footnote{http://rev.ai} API.

Today, we are open-sourcing our core speech recognition and diarization models for non-commercial use. We are releasing both a full production pipeline for developers as well as pared-down research models for experimentation. Rev hopes that these releases will spur research and innovation in the fast-moving domain of voice technology. The speech recognition models released today outperform all existing open source speech recognition models across a variety of long-form speech recognition domains.

This release, which we are calling Reverb, encompasses two separate models: an automatic speech recognition (ASR) model in the WeNet\footnote{https://github.com/wenet-e2e/wenet} framework and two speaker diarization models in the Pyannote\footnote{https://github.com/pyannote/pyannote-audio} framework. For researchers, we provide simple scripts for combining ASR (``Reverb Research") and diarization output into a single diarized transcript. For developers, we provide a full pipeline (``Reverb") that handles both ASR and diarization in a production environment. Additionally, we are releasing an int8 quantized version of the ASR model within the developer pipeline (“Reverb Turbo”) for applications that are particularly sensitive to speed and/or memory usage.

Reverb ASR was trained on 200,000 hours of English speech, all expertly transcribed by humans --- the largest corpus of human transcribed audio ever used to train an open-source model. The quality of this data has produced the world’s most accurate English ASR system, using an efficient model architecture that can be run on either CPU or GPU. 

Additionally, this model provides user control over the level of verbatimicity of the output transcript, making it ideal for both clean, readable transcription and use-cases like audio editing that require transcription of every spoken word including hesitations and re-wordings. Users can specify fully verbatim, fully non-verbatim, or anywhere in between for their transcription output. 

For diarization, Rev used the high-performance pyannote.audio\footnote{https://github.com/pyannote/pyannote-audio} library to fine-tune existing models on 26,000 hours of expertly labeled data, significantly improving their performance. Reverb diarization v1 uses the pyannote3.0\footnote{https://huggingface.co/pyannote/speaker-diarization-3.0} architecture, while Reverb diarization v2 uses WavLM \cite{Chen_2022} instead of SincNet \cite{ravanelli2019speakerrecognitionrawwaveform} features. 

\section{Reverb ASR}
\subsection{Data}

Rev’s ASR dataset is made up of long-form, multi-speaker audio featuring a wide range of domains, accents and recording conditions. This corpus contains audio transcribed in two different styles: verbatim and non-verbatim. 

Verbatim transcripts include all speech sounds in the audio (including false starts, filler words, and laughter), while non-verbatim transcripts have been lightly edited for readability; see \Cref{verbatimicity_examples} for specific examples. Training on both of these transcription styles is what enables the style control feature of the Reverb ASR model.

To prepare our data for training, we employ a joint normalization and forced-alignment process, which allows us to simultaneously filter out poorly-aligned data and get the best possible timings for segmenting the remaining audio into shorter training segments. During the segmentation process, we include multi-speaker segments, so that the resulting ASR model is able to effectively recognize speech across speaker switches. 

The processed ASR training corpus comprises 120,000 hours of speech with verbatim transcription labels and 80,000 hours with non-verbatim labels.

\subsection{Model Architecture}

Reverb ASR was trained using a modified version of the WeNet toolkit and uses a joint CTC/ attention architecture\footnote{https://www.rev.com/blog/speech-to-text-technology/what-makes-revs-v2-best-in-class}. The encoder has 18 conformer layers and the bidirectional attention decoder has 6 transformer layers, 3 in each direction. In total, the model has approximately 280M parameters. 

One important modification available in Rev’s WeNet release is the use of the language-specific layer mechanism \cite{lsl}. While this technique was originally developed to give control over the output language of multilingual models, Reverb ASR uses these extra weights for control over the verbatimicity of the output. These layers are added to the first and last blocks of both the encoder and decoder. 

The joint CTC/ attention architecture enables experimentation with a variety of inference modes, including: 
greedy CTC decoding, CTC prefix beam search (with or without attention rescoring), attention decoding, and joint 
CTC/ attention decoding. The joint decoding available in Rev’s WeNet is a slightly modified version of the time synchronous joint decoding implementation from ESPnet\footnote{https://github.com/espnet/espnet}. 

The production pipeline uses WFST-based beam search with a simple unigram language model on top of the encoder outputs, followed by attention rescoring. This pipeline also implements parallel processing and overlap decoding at multiple levels to achieve the best possible turn-around time without introducing errors at the chunk boundaries. While the research model outputs unformatted text, the production pipeline includes a post-processing system for generating fully formatted output. 

\subsection{Benchmarks}

Unlike many ASR providers, Rev primarily uses long-form speech recognition corpora for benchmarking. We use each model to produce a transcript of an entire audio file, then use \texttt{fstalign}\footnote{https://github.com/revdotcom/fstalign/} to align and score the complete transcript. We report micro-average WER across all of the reference words in a given test suite. As part of our model release, we have included our scoring scripts so that anyone can replicate our work, benchmark other models, or experiment with new long-form test suites. 

Here, we’ve benchmarked Reverb ASR against the best performing open-source models currently available: OpenAI’s Whisper large-v3\footnote{https://huggingface.co/openai/whisper-large-v3} and NVIDIA’s Canary-1B\footnote{https://huggingface.co/nvidia/canary-1b}. Note that both of these models have significantly more parameters than Reverb ASR. For these models and Rev’s research model, we use simple chunking with no overlap --- 30s chunks for Whisper and Canary, and 20s chunks for Reverb. The Reverb research results use CTC prefix beam search with attention rescoring. We used Canary through Hugging Face and used the WhisperX\footnote{https://github.com/m-bain/whisperX} implementation of Whisper. For Whisper, we use NeMo\footnote{https://github.com/NVIDIA/NeMo-text-processing} to normalize the model outputs before scoring.

For long-form ASR, we’ve used three corpora: 
Rev16 \cite{radford2023robust} (podcasts), 
Earnings21 \cite{Del_Rio_2021} (earnings calls from US-based 
companies), and Earnings22 \cite{earnings22} (earnings calls from 
global companies); results are shown in \Cref{tab:Earnings}.

\begin{table}[]
\centering
\caption{WER on the Earnings21 and Earnings22 test sets.}
\vspace{0.5em}
\begin{tabular}{l|l|l}
 Model & Earnings21 & Earnings22 \\
 \hline
 Reverb Verbatim & 7.64 & 11.38 \\
 Reverb Turbo Verbatim & 7.88 & 11.60\\
 Reverb Research Verbatim & 9.68 & 13.68 \\
 Whisper large-v3 & 13.67 & 18.53 \\
 Canary-1B & 14.40 & 19.01
\end{tabular}
\label{tab:Earnings}
\end{table}
\vspace{0.5em}

For Rev16, we have produced both verbatim and non-verbatim human transcripts. For all Reverb models, we run in verbatim mode for evaluation with the verbatim reference and non-verbatim mode for evaluation with the non-verbatim reference; results are shown in \Cref{tab:Rev16}.

\begin{table}[]
\centering
\caption{WER on the Rev16 test set, comparison between verbatim references and non-verbatim references. }
\vspace{0.5em}
\begin{tabular}{l|l|l}
 Model & Verbatim Reference & Non-Verbatim Reference \\
 \hline
 Reverb & 7.99 & 7.06 \\
 Reverb Turbo & 8.25 & 7.50\\
 Reverb Research & 10.30 & 9.08 \\
 Whisper large-v3 & 10.67 & 11.37 \\
 Canary-1B & 13.82 & 13.24
\end{tabular}
\label{tab:Rev16}
\end{table}
\vspace{0.5em}

We have also used GigaSpeech \cite{GigaSpeech2021} for a non-Rev transcribed benchmark. We ran Reverb ASR in verbatim mode and used the HuggingFace Open ASR Leaderboard\footnote{https://huggingface.co/spaces/hf-audio/open\_asr\_leaderboard} evaluation scripts; results are shown in \Cref{tab:GigaSpeech}. 

\begin{table}[]
\centering
\caption{WER on the GigaSpeech test set. }
\vspace{0.5em}
\begin{tabular}{l|l}
 Model & GigaSpeech \\
 \hline
 Reverb Research Verbatim & 11.05 \\
 Whisper large-v3 & 10.02 \\
 Canary-1B & 10.12
\end{tabular}
\label{tab:GigaSpeech}
\end{table}
\vspace{0.5em}

Overall, Reverb ASR significantly outperforms the competition on long-form ASR test suites. Rev’s models are particularly strong on the Earnings22 test suite, which contains mainly speech from non-native speakers of English. We see a small WER degradation from the use of the Turbo model, but a much larger gap between the production pipeline and research model - demonstrating the importance of engineering a complete system for long-form speech recognition. 

On the GigaSpeech test suite, Rev’s research model is worse than other open-source models. The average segment length of this corpus is 5.7 seconds; these short segments are not a good match for the design of Rev’s model. These results demonstrate that despite its strong performance on long-form tests, Rev may not be not the best candidate for short-form recognition applications like voice search. 

\subsection{Verbatimicity}
\label{verbatimicity_examples}

Rev has the only AI transcription API and model that allows user control over the verbatimicity of the output. The developer pipeline offers a verbatim mode that transcribes all spoken content and a non-verbatim mode that removes unnecessary phrases to improve readability. The output of the research model can be controlled with a verbatimicity parameter that can be anywhere between zero and one. 

The Rev team has found that halfway between verbatim and non-verbatim produces a reader-preferred style for captioning --- capturing all content while reducing some hesitations and stutters to make captions fit better on screen. \Cref{tab:verbatimstyles} illustrates a range of verbatim expressions and how they tend to be treated by different models. \Cref{tab:GigaSpeechVerbatimicity} illustrates the three Reverb verbatimicity levels with respect to a GigaSpeech utterance.

\begin{table}[]
\centering
\caption{Verbatim styles}
\vspace{0.5em}
\begin{tabular}{l|c|c}
Verbatim expression & Other models transcribe & Reverb verbatim transcribes \\
\hline
Repeated stutter words &  & \checkmark \\
Repeated phrases & & \checkmark \\
Filled pauses (um, uh) & & \checkmark \\
``you know” & \checkmark & \checkmark \\
``kind of"  & \checkmark & \checkmark \\
``sort of"  & \checkmark & \checkmark \\
``like"  & \checkmark & \checkmark \\
\end{tabular}
\label{tab:verbatimstyles}
\end{table}
\vspace{0.5em}

\begin{table}[]
\centering
\caption{Example using GigaSpeech \texttt{POD1000000032\_S0000058}}
\vspace{0.5em}
\begin{tabular}{l|p{0.6\linewidth}}
Style & Transcription \\
\hline
Reverb verbatim &
and and if you if you try and understand which ones there are you it's it's a it's a long list \\
Reverb non-verbatim & 
and if you try and understand which ones there are it's a long list \\

Reverb half-verbatim & 
and if you if you try and understand which ones there are you it's a long list \\
\end{tabular}
\label{tab:GigaSpeechVerbatimicity}
\end{table}

\section{Reverb Diarization}

\subsection{Data}
Rev’s diarization training data comes from the same diverse corpus as the ASR training data. However, annotation for diarization is particularly challenging, because of the need for precise timings, specifically at speaker switches, and the difficulties of handling overlapped speech. As a result, only a subset of the ASR training data is usable for diarization. The total corpus used for diarization is 26,000 hours.

\subsection{Model Architecture}
The Reverb diarization models were developed using the pyannote.audio library \cite{Plaquet23,Bredin23}. Reverb diarization v1 is identical to pyannote3.0\footnote{https://huggingface.co/pyannote/speaker-diarization-3.0} in terms of architecture but it is fine-tuned on Rev’s transcriptions for 17 epochs. Training took 4 days on a single A100 GPU. The network has 2 LSTM layers with hidden size of 256, totaling approximately 2.2M parameters. Our most precise diarization model - Reverb diarization v2 - uses WavLM instead of the SincNet features in the pyannote3.0 basic model.

\subsection{Benchmarks}

While DER is a valuable metric for assessing the technical performance of a diarization model in isolation, WDER \cite{shafey2019jointspeechrecognitionspeaker} (Word Diarization Error Rate) is more crucial in the context of ASR because it reflects the combined effectiveness of both the diarization and ASR components in producing accurate, speaker-attributed text. In practical applications where the accuracy of both “who spoke” and “what was spoken” is essential, WDER provides a more meaningful and relevant measure for evaluating system performance and guiding improvements. For this reason we only report WDER metrics. We show results for two test suites, Earnings21 and Rev16, in \Cref{tab:diarizationresults}.

. 

\begin{table}[]
\centering
\caption{WDER on the Earnings21 and Rev16 test sets using Reverb }
\vspace{0.5em}
\begin{tabular}{l | l | l}
 Model & Earnings21 & Rev16 \\
 \hline
 Pyannote3.0 & 0.051 & 0.090 \\
 Reverb Diarization v1 & 0.047 & 0.077 \\
 Reverb Diarization v2 & 0.046 & 0.078 \\
\end{tabular}
\label{tab:diarizationresults}
\end{table} 
\vspace{0.5em}

\section{Conclusion}
Rev is excited to release the state-of-the-art Reverb ASR and diarization models to the public. We hope that these releases will spur research and innovation in the fast-moving domain of voice technology. To get started, visit https://github.com/revdotcom/reverb for research models or https://github.com \linebreak 
/revdotcom/revai for the complete developer solution. Schedule a demo today\footnote{https://www.rev.com/online-transcription-services/contact-sales} to learn more about the Rev.ai API or commercial licensing for Reverb. 

%
%
\bibliographystyle{splncs04}
\bibliography{biblio}
\end{document}